# Unravelling Causal Genetic Biomarkers of Alzheimer's Disease via Neuron to Gene-token Backtracking in Neural Architecture: A Groundbreaking Reverse-Gene-Finder Approach

**Victor OK Li[1*], Yang Han[1], Jacqueline CK Lam[1*]**

[1]Department of Electrical and Electronic Engineering, The University of Hong Kong, Hong Kong
{vli, yhan, jcklam}@eee.hku.hk

**Abstract**

Alzheimer's Disease (AD) affects over 55 million people globally, yet the key genetic contributors remain poorly understood. Leveraging recent advancements in genomic foundation models, we present the innovative Reverse-Gene-Finder technology, a ground-breaking neuron-to-gene-token backtracking approach in a neural network architecture to elucidate the *novel causal genetic biomarkers* driving AD onset. Reverse-Gene-Finder comprises three key innovations. Firstly, we exploit the observation that genes with the highest probability of causing AD, defined as the most causal genes (MCGs), must have the highest probability of activating those neurons with the highest probability of causing AD, defined as the most causal neurons (MCNs). Secondly, we utilize a gene token representation at the input layer to allow each gene (known or novel to AD) to be represented as a discrete and unique entity in the input space. Lastly, in contrast to the existing neural network architectures, which track neuron activations from the input layer to the output layer in a feed-forward manner, we develop an innovative backtracking method to track backwards from the MCNs to the input layer, identifying the Most Causal Tokens (MCTs) and the corresponding MCGs. Reverse-Gene-Finder is highly interpretable, generalizable, and adaptable, providing a promising avenue for application in other disease scenarios.

**Code** — https://github.com/yanghangit/RGF/

## 1. Introduction

Alzheimer's disease (AD) is a progressive neurodegenerative disorder that represents the most common cause of dementia worldwide. It affects over 55 million people globally. The societal and economic impact of AD is profound, with costs associated with care and treatment exceeding hundreds of billions of dollars annually, placing an immense burden on families and healthcare systems (Alzheimer's Association 2024).

Genetic studies over the past decades have significantly advanced our understanding of AD, uncovering genetic biomarkers associated with AD and providing valuable insights into the biological pathways involved in the disease. However, the complex etiology of AD remains not fully understood, highlighting the need to develop innovative methods to unravel the causal genetic biomarkers of AD and deepen our understanding of the disease mechanisms. The rest of this section provides (1) an overview of AD, and the key known genes associated with the disease, (2) a review of recent advances in deep learning techniques for AD genetic biomarker identification to highlight the research gaps in existing gene-finding approaches, and (3) the significance and objectives of this study.

### 1.1 Brief Overview of Alzheimer's Disease and Key Genes Characterizing AD

AD is characterized by the accumulation of amyloid-beta plaques and neurofibrillary tangles in the brain, leading to cognitive decline, memory loss, and, ultimately, loss of independent function. Understanding the genetic underpinnings of AD has been a central focus of research in the field. One of the most prominent genetic biomarkers associated with AD is the APOE gene (Farrer et al. 1997). Individuals who carry one or two copies of the APOE ε4 allele often (but not necessarily) have a significantly higher risk of developing AD. In addition to APOE ε4, genome-wide association studies (GWAS) have identified other genetic variants that contribute to a modest increase in risk for AD, such as those in BIN1 and CLU genes (Bertram et al. 2007). Despite these findings, the limitations of GWAS make it challenging to identify the causal AD biomarkers from the known AD-associated biomarkers (Neuner, Julia, and Goate 2020).

The pathogenesis of AD is highly complex, involving interactions across different genes, somatic mutations, epigenetic modifications, and influences on genetic expressions due to environmental factors (Downey et al. 2022; Migliore and Coppede 2022). This complexity poses significant challenges in predicting and understanding the etiology and progression of AD based solely on a list of correlated genetic biomarkers. The complexity highlights the need for a more sophisticated approach via identifying the causal genetic biomarkers of AD and unravelling their underlying causal mechanisms.

**1.2 Related Work and Research Gaps of Existing Gene-Finding Approaches**

Recent advancements in artificial intelligence (AI) techniques, particularly deep learning, offer a promising opportunity to integrate vast amounts of genetic and clinical data. These novel techniques carry the potential to identify new genetic biomarkers, uncover hidden patterns, and provide a more complete understanding of the genetic architecture of AD, ultimately leading to better predictive models and therapeutic targets.

Several deep learning models have been developed to identify genetic variants of AD and related neuro-diseases, outperforming traditional statistical approaches. For example, a three-step approach was proposed for AD genetic biomarker identification: (1) using a convolutional neural network model to select phenotype-associated fragments from the whole genome, (2) identifying informative single nucleotide polymorphisms (SNPs) from the selected fragments based on phenotype influence scores, and (3) developing a classification model based on the identified SNPs (Jo et al. 2022). An interpretable neural network model with hierarchical layers has been developed to reduce the number of weight parameters and activations and improve computational efficiency, using multiple knockoffs in the input layer to rigorously control the false discovery rate when identifying genetic biomarkers for AD using feature importance analysis (Kassani et al. 2022). A hierarchical graph attention network has also been developed to identify genetic biomarkers for post-traumatic stress disorder via saliency analysis (Zhang et al. 2023).

Moreover, increasing deep learning studies have focused on the transcriptomic and epigenetic changes associated with AD. For example, deep neural network models were developed to predict AD using gene expression and DNA methylation data (Lee and Lee 2020; Park, Ha and Park 2020). Further, recent advances in single-cell genetic sequencing have advanced our understanding of the molecular and cellular architectures of AD (Mathys et al. 2019; Rahimzadeh et al. 2024). Deep learning-based AD studies have investigated missing gene expression data imputation and cell clustering and the identification of significant genes associated with AD using single-cell gene expression data (Wang et al. 2021, 2022). However, existing data-driven gene-finding approaches often operate on a high-dimensional genetic search space and are constrained in performance due to small datasets, making it extremely challenging to identify causal genetic biomarkers.

Recent advancements in large foundation models pre-trained on large-scale genomics datasets have opened new avenues for understanding the complex interplay of disease-associated genes. These models, such as Geneformer (Theodoris et al. 2023), often utilize the transformer architecture pre-trained on vast amounts of data, while leveraging its self-attention mechanisms to capture intricate genetic interactions in network biology, to enhance the performance of downstream tasks such as disease prediction, even in cases of limited data availability.

Although these foundation models have shown impressive performance in applications such as disease prediction, they are black-box models, presenting significant challenges in understanding and interpreting model results. Previous studies have attempted to uncover these black-box models using explainable and interpretable AIs. These methods can be considered as local analyses, which seek to understand how specific predictions are generated by the models, e.g., using gradient-based feature attribution methods and global analysis to uncover the inner mechanics of the model, investigating how individual neurons and their interconnections work (Luo and Specia 2024).

In particular, the causal tracing technique (Meng et al. 2022) has been developed to study the causal impacts of neuron activations on large language model (LLM) outputs. Specifically, this technique represents the internal computation of an LLM as a grid of hidden states and the relationships between these states. It takes the grid as a causal graph and uses causal mediation analysis to measure the impact of intermediate variables in the graph by observing the graph's internal activations during various runs, including (1) a clean run for prediction as usual, (2) a corrupted run where the prediction is damaged, and (3) a restore run that investigates how one hidden state can recover the damaged prediction. This causal mediation analysis has made it possible to assess the contribution of each hidden state in predicting factual statements, providing new insights into the computational mechanisms of LLMs.

However, how large foundation models can be utilized to identify causal genetic biomarkers of AD has yet to be fully understood. Identifying the causal genetic biomarkers of the disease and interpreting the results based on black-box foundation models remains challenging. To the best of our knowledge, this study is the first to use a causal tracing technique to interpret an AD-specific large genomic black-box foundation model, and the first to develop a Reverse-Gene-Finder, a novel neuro-to-gene token backtracking method to identify the potentially causal novel[1] genetic candidates that contribute to AD.

### 1.3 Research Significance and Objectives

This study introduces a groundbreaking methodology to identify novel causal genetic biomarkers for AD, focusing on uncovering the underlying genetic causes of the disease instead of improving prediction performance. Our innovative Reverse-Gene-Finder technology represents a significant advancement in AD research, offering an interpretable framework to identify novel genes that may play a crucial role in the etiology of AD. Unlike traditional studies that prioritize prediction accuracy or the identification of associated genetic markers, our approach directly addresses the critical challenge of causal gene discovery. The innovative approach proposed in this research offers substantial improvements over existing gene-finding studies focused on identifying genetic biomarkers for AD, addressing critical challenges in interpreting and discovering causal genetic biomarkers. The findings of this study underscore the importance of innovative methodologies in advancing our understanding of complex diseases like AD. Our proposed reverse engineering approach reveals the novel causal genetic biomarkers of AD and opens new avenues for exploring genetic causes driving other diseases, potentially transforming the landscape of genomic research and precision medicine.

The objectives of this study are three-fold. Firstly, this study aims to harness the immense potential of large foundation models by utilizing genetic data from individuals diagnosed with AD, as well as healthy controls, to fine-tune a pre-trained genomic foundation model. This model is optimized explicitly for classifying AD, thereby enabling the extraction of meaningful insights from complex and high-dimensional genomic data.

Secondly, this study aims to systematically identify the most causal neurons (MCNs) within the graph of the fine-tuned genomic foundation model, which carries the most significant causal effects for the prediction of AD onset. By modifying the input data to mask out the known genes strongly associated with AD, this study aims to identify MCNs to enable backtracking and improve model interpretability.

Finally, this study aims to backtrack from the MCNs to the input layer to identify the most causal tokens (MCTs) and the most causal genes (MCGs) they represent. We exploit the observation that genes with the highest probability of causing AD, defined as MCGs, must have the highest probability of activating those neurons with the highest probability of causing AD, defined as MCNs. This reverse engineering approach can lead to the discovery of novel genetic biomarkers for AD. The insights gained from this process will deepen our understanding of the genetic basis of AD and hold significant potential for application in different disease contexts, offering a versatile framework for uncovering genetic biomarkers and causal factors in a wide range of classification tasks.

## 2. Methodology

### 2.1 The Reverse-Gene-Finder Mechanism

Figure 1 summarizes the overarching Reverse-Gene-Finder mechanism. Firstly, we fine-tune a genomic foundation model specifically for AD classification, utilizing single-cell gene expression data from AD patients and healthy control subjects. Secondly, by modifying the input data by masking out known genes strongly associated with AD, we systematically identify MCNs related to AD, employing causal tracing techniques to observe the effects of in-silico perturbations of such known genes on neurons across different layers of the fine-tuned Gene-former. Finally, our Reverse-Gene-Finder backtracking method enables the identification of MCTs (and the MCGs they represent) most likely to activate MCNs.

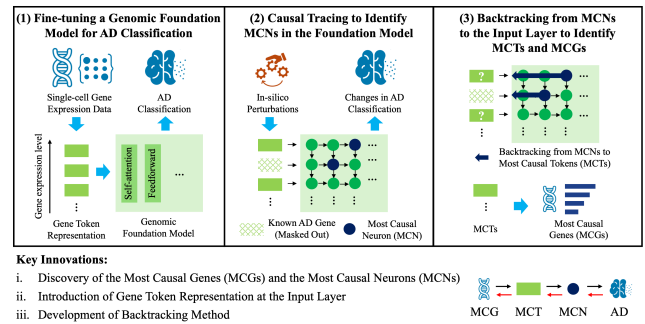

Figure 1. Overarching Reverse-Gene-Finder Mechanism and Three Novelties

### 2.2 Novelties

Our novel Reverse-Gene-Finder technology comprises three key innovations (see Figure 1). The details are listed below.

### 2.2.1 Discovery of the Most Causal Genes (MCGs) and the Most Causal Neurons (MCNs)

We exploit the observation that genes with the highest probability of causing AD, defined as the MCGs, must have the highest probability of activating those neurons with the highest probability of causing AD, defined as the MCNs.

---

[1] "Novel" refers to the identification of new genetic candidates that are not in the list of previously known (model inputted) AD-associated genes.

How can we locate the MCNs with the highest probability of causing AD? This study adopts the causal tracing technique initially developed for LLMs (Meng et al. 2022) to identify the MCNs in a genomic foundation model with the following assumptions and modifications. We assume that when key information about the AD-associated genes (obtained from previous literature on AD genetic association studies) in the input data is masked (or perturbed), the probability of AD prediction will likely change, making it possible to observe the effects of in-silico perturbations of known AD genes on disease onset prediction, which are mediated by neurons in the genomic foundation model.

Based on this assumption, the difference between the probability of getting AD in (1) the baseline corrupted run (where AD-associated genes are corrupted by noise) and (2) the corrupted-with-restoration run (where AD-associated genes remain corrupted, but a particular neuron (i.e., hidden state is reverted to the clean version) is calculated to measure the indirect effect of the hidden state in the model. The indirect effects of hidden states across different layers are then used to quantify the importance of different hidden states and select the MCNs in predicting AD.

### 2.2.2 Introduction of Gene Token Representation at the Input Layer

We utilize a gene token representation at the input layer, where genes are represented by gene tokens, so that when we backtrack from the MCNs to the input layer, we can identify the MCTs most likely to cause AD. The gene token representation in the input layer allows each gene (known or novel to AD) to be represented as a discrete and unique entity in the input space, such that we can discover new gene tokens corresponding to novel causal AD genes, previously not being known as causal genetic biomarkers of AD.

Specifically, each input record consists of a list of genes. Each gene has a measurement value (e.g., gene expression level) and is assigned to a unique gene token ID. The genes at the input layer are then represented by a vector of gene tokens: $\{x_1, x_2, ..., x_T\}$. The position of each gene token in the input vector is determined by the measurement value, e.g., using rank value encoding (Theodoris et al. 2023).

While only a small proportion of the gene tokens correspond to the known AD genes, using a gene token representation can incorporate other gene tokens in the input space. Some of these gene tokens may correspond to novel genes (that are previously unknown to the model) contributing to AD, potentially offering new insights into the genetic mechanism of AD.

### 2.2.3 Development of Backtracking Method

In contrast to the existing neural network architectures, which track neuron activations from the input layer to the output layer in a feed-forward manner, we develop an innovative backtracking method to track backwards from the MCNs to the input layer, identifying the MCTs and, consequently, the corresponding MCGs. By backtracking via computing scores iteratively from the output back to the input, this novel method generates a cumulative measure of influence score for each gene token, highlighting the gene tokens most likely to activate MCNs.

The information flowing from the input layer to the MCNs are backtracked and summarized to identify the MCTs more likely to activate these MCNs. Assume the neural network model has $L$ layers. The $i$th hidden state at layer $l$ is denoted as $h_i^l$. The identified MCNs are denoted as $\{\hat{h}_i^l\}$. The activation (i.e., indirect causal effect) of each MCN $\hat{h}_i^l$ is denoted as IE$(\hat{h}_i^l)$. For each input, the score of a gene token at position $i$ is quantified by the weighted sum of the accumulated indirect effect of all MCNs backtracked to the $i$th unit in the input layer (see Algorithm 1, which shows how to perform backtracking in an iterative manner). Specifically, the score of the $i$th unit at layer $l$, denoted as $s_i^l$, is a weighted sum of all interconnected MCNs in the next layer $l + 1$, where each MCN $k$ is quantified by the indirect effect IE$(\hat{h}_k^{l+1})$ and the score $s_k^{l+1}$ at layer $l + 1$. This score is computed iteratively from the last layer to the input layer. The weights $W$ measure the strength of interconnectivity between one hidden state and another across layers. It can be determined by the fitted model parameters, e.g., self-attention weights in transformers. Gene token scores are calculated and averaged after iterating every position in each input sample. The gene tokens with the highest scores are identified as MCTs. These gene tokens can be either known or novel (unknown to the model) genetic biomarkers for AD. The MCGs corresponding to the identified MCTs are selected as the putative causal genetic biomarkers for AD.

---

**Algorithm 1: Reverse-Gene-Finder Backtracking Method**

**Input**: Input Position $i$, MCNs $\{\hat{h}_i^l\}$, Model Weights $W$
**Parameter**: Number of Layers $L$
**Output**: Gene Token Score at Position $i$
1: Initialize $s_i^L = 0$
2: **for** $l$ from $L - 1$ **to** $1$
3: $\quad s_i^l = \sum_k W_{i,k}^l * (\text{IE}(\hat{h}_k^{l+1}) + s_k^{l+1})$
4: **end for**
5: **return** $s_i^1$

---

## 3. Experimental Setting

This section details the experimental setup, using a single-cell genetic dataset from a comprehensive AD case/control study (Sun et al. 2023) and a large genomic foundation model (namely, Geneformer (Theodoris et al. 2023)) to demonstrate the proposed Reverse-Gene-Finder technology.

All experiments were conducted using a Nvidia A100 40GB GPU on a Linux system via Google Colab, with Python (version 3.10.12) and deep learning libraries PyTorch (version 2.3.1+cu121) and Transformers (version 4.44.0). The pre-trained 12-layer Geneformer model was obtained from HuggingFace[2]. The causal tracing analysis was developed based on the code from GitHub[3]. Pathway enrichment analysis was performed using STRING (version 11.0)[4].

## 3.1 Fine-Tuning a Genomic Foundation Model for AD Classification

### 3.1.1 Use of Single-Cell Gene Expression Data from AD Patients and Healthy Control Subjects

Since the genomic foundation model used this study was pre-trained on single-cell gene expression data (which can be seen as a large sparse matrix of raw count values by cells and genes), this study utilized a publicly available single-cell gene expression dataset based on the Religious Orders Study and Memory and Aging Project (ROSMAP), a large comprehensive longitudinal study involving AD patients and healthy control subjects, integrating clinical and genetic data, with single-cell gene expression data available (Bennett et al. 2018). Specifically, single-cell microglial transcriptomic data collected from 443 human subjects (postmortem brain samples) in the ROSMAP study, labeled non-AD, early-AD, and late-AD, were obtained from the supplementary website[5] provided by Sun et al. (2023).

Although DNA is stable, gene expression levels may vary over time along with the aging process (Srivastava et al. 2020). Since this study focuses on early-stage AD genetic biomarkers, which are more likely to be upstream biomarkers leading to AD, 132 early-AD and 219 non-AD subjects were selected from the dataset. Single-cell samples were grouped into 12 clusters as distinct microglial states, and the 10th cluster (one of the inflammatory states) from the prefrontal cortex region was selected because it showed more significant transcriptional changes in early AD stages compared to other microglial states (Sun et al. 2023). Standard quality controls were performed by filtering out noisy cells that (1) had unique feature counts greater than 2,500 or less than 200 or (2) had more than 5% mitochondrial counts.

After data preprocessing, the single-cell samples were further divided into five folds for cross-validation, using stratified sampling based on diagnosis labels to ensure a balanced representation of single-cell samples across different folds. For each trial of the five-fold cross-validation, 80% of the samples was used for training and validation, and the remaining 20% was used for testing. The number of training, validation, and testing samples in each cross-validation trial was 489, 123, and 153, respectively.

### 3.1.2 Representation of Genes as Gene Tokens

After data preprocessing, the total number of genes was 15,549. The rank value encoding method (Theodoris et al. 2023) was adopted to represent the input genes. Specifically, each input consisted of gene expression levels measured in a single cell. Each gene was assigned to a unique gene token ID based on the token dictionary provided by the pre-trained genomic model. In each input vector, gene tokens were ranked based on their corresponding gene expression levels.

### 3.1.3 Fine-tuning for AD classification

A binary classification layer was added to the pre-trained Geneformer model for predicting AD status, allowing it to adapt to the specific characteristics of AD gene expression patterns. The model input was the single-cell gene token representation, and the output was the state of the input cell (i.e., non-AD or early-AD). The training objective was a cross-entropy loss function based on the predicted and ground truth labels. The best hyperparameters were chosen based on the validation performance. Using the optimized hyperparameters, the preprocessed data were used to fine-tune the final Geneformer model for AD classification.

Specifically, the 12-layer pre-trained Geneformer model was used in this study. The first four layers (i.e., one-third of all layers) were frozen during fine-tuning. Hyperparameters were optimized using HyperOpt (Liaw et al. 2018). The best hyperparameters were selected based on the performance of the validation set using the F1-score, which combines precision and recall into a single metric. The hyperparameter ranges and the selected ones are detailed in the code.

The performance of the final fine-tuned AD classification model was evaluated by the Area under the ROC Curve (AUC) score, which comprehensively measures how much the fine-tuned model is capable of distinguishing between early-AD and non-AD cells across varying classification thresholds. Moreover, the performance was evaluated by sensitivity (i.e., true positive rate) and specificity (i.e., true negative rate), which are often used in the context of medical diagnosis. All performance evaluation results were reported by their mean and standard deviation using five-fold cross-validation (i.e., the number of runs was five).

## 3.2 Identification of MCNs Related to AD

### 3.2.1 Known Genes Related to AD

The genes known to be highly associated with AD were obtained from the AlzGene database, curated based on systematic meta-analyses of AD genetic association studies (Bertram et al. 2007). Specifically, the following top AD-associated genes were selected from http://www.alzgene.org/TopResults.asp, including APOE, BIN1, CLU, ABCA7, CR1, PICALM, MS4A6A, CD33, MS4A4E, and CD2AP.

---

[2] https://huggingface.co/ctheodoris/Geneformer
[3] https://github.com/kmeng01/rome/
[4] https://string-db.org/help/api/
[5] https://compbio.mit.edu/microglia_states/

In our initial masking process, we only masked genes that are universally recognized as closely associated with AD. For example, APOE was used to evaluate Reverse-Gene-Finder's ability to uncover novel genetic biomarkers. Many genes, including MT-CO1, DOCK4, and RUNX1, though being associated with AD in specific contexts (as evident in previous literature), are not recognized universally as AD genes. Hence, they were not being masked. Once these genes are verified as universally recognized AD genes, we will input these genes in the later masking process.

### 3.2.1 Use of Causal Tracing Analysis

A causal tracing analysis was performed using all single-cell samples in the testing set across five-fold cross-validation. In-silico perturbations were performed by adding a zero-mean Gaussian noise to the original gene token embeddings corresponding to AD-associated genes. The noise level was set to 1. The maximum number of genes in one input was set to 256 to reduce computational costs. For an input vector exceeding this threshold, the top and bottom tokens (i.e., the most upregulated and downregulated genes) were kept; otherwise, padding tokens were added. The hidden states in the Geneformer with the highest indirect causal effects (based on the 95th percentile cutoff) were selected as MCNs.

### 3.3 Identification of MCTs and MCGs via Reverse-Gene-Finder Backtracking

Since Geneformer is a transformer-based foundation model, the weights used in the backtracking method were self-attention weights $A_{i,j}^l$ at each layer $l$, measuring the relative importance between one MCN $i$ and another MCN $j$. The gene token score at every input position was calculated for each input sample based on the weighted sum of MCN activations from the last layer backtracked to the input layer (see Algorithm 1). Some of these gene tokens corresponded to previously unknown genetic candidates for AD. The top 10 MCTs were identified based on the gene token scores averaged across all input samples. The top 10 MCGs corresponding to the top 10 MCTs were further identified.

### 3.4 Validation and Pathway Enrichment Analysis

The identified MCGs were validated based on cross-referencing previous literature related to AD. Moreover, pathway enrichment analysis was performed to reveal the relevant biological pathways involved in AD based on the identified MCGs. Specifically, the STRING analysis identified pathways enriched with the identified MCGs against randomized gene sets (Szklarczyk et al. 2019). Since multiple pathways were tested simultaneously, multiple testing correction was adopted using the false discovery rate (FDR). Statistically significant pathways from the KEGG pathway database[6] with an FDR of less than 0.05 were obtained.

---
[6] https://www.genome.jp/kegg/

## 4. Results and Discussion

### 4.1 Discovery of Previously Unknown Genetic Candidates

The AUC score of the fine-tuned Geneformer model based on five-fold validation is 74.67% ± 6.74%. The sensitivity and specificity scores are 62.68% ± 11.43% and 73.22% ± 16.01%, respectively. It should be noted that the focus of our study is more about demonstrating a new approach that can help us uncover novel causal genes via backtracking than the fine-turned model performance. Nevertheless, this AUC score is reasonable given the single use of genetic data and is comparable to the AUC score reported in previous AD prediction models trained on genetic data (Escott-Price et al. 2017). Further improvement in performance is expected after incorporating more data modalities, such as proteomic and brain imaging data.

Figure 2 shows the locations of MCNs with the strongest average indirect effects based on the 95th percentile cutoff. MCNs are more likely to be in the front and middle layers. Some MCNs are at the top or the bottom positions, which is not surprising given the rank encoding of input data, where over-expressed and under-expressed genes are located at the top and bottom of the input vector, respectively. Other MCNs are closer to the middle positions, suggesting that the importance of genes is not only determined by their expression levels but also by how they interact with other genes.

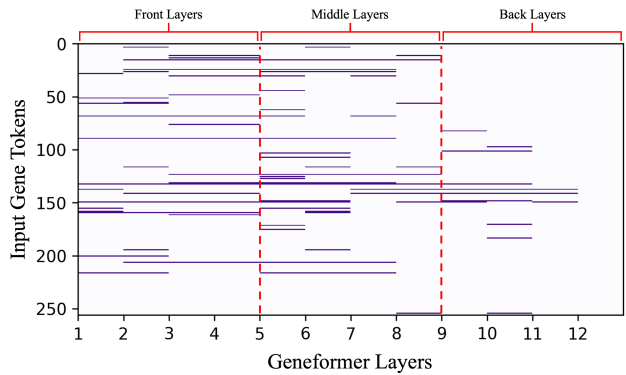

Figure 2: Locations of the Most Causal Neurons (Based on the 95th Percentile Cutoff).

The top 10 MCGs (with scores) based on backtracking from the top MCNs are PLXDC2 (0.499), MT-CO3 (0.498), DOCK4 (0.482), ARHGAP24 (0.450), MEF2A (0.428), RUNX1 (0.427), ITPR2 (0.426), FOXN3 (0.406), MT-CO1 (0.392), and SORL1 (0.389). These top MCGs are all novel genes not included in the known AD gene list (which have previously been inputted to the Geneformer model to locate MCNs before backtracking; see Section 3.2). In other words, the novel Reverse-Gene-Finder carries the potential to unravel new genes and pathways driving AD onset.

## 4.2 Validation of Discovered Genetic Candidates via Literature Review

Each identified MCG has been further validated using previous literature related to AD. PLXDC2 was linked to the progression of both amyloid beta and tau, the two hallmarks of AD (Bartolo et al. 2022). The mitochondrial genes MT-CO3 and MT-CO1 are crucial for cellular energy production, and their dysfunction was associated with the pathogenesis of AD due to the increased oxidative stress in human neurons (Lunnon et al. 2017). The DOCK family, including DOCK4, was associated with several neurodegenerative and neuropsychiatric diseases, including AD and autism spectrum disorders (Shi, 2013). ARHGAP24 was identified as a novel gene in mild cognitive impairment (MCI) subjects from a GWAS study (Chung et al. 2018). MEF2A, a transcription factor in regulating neural transmission and synaptic density, could play a key role in cognitive resilience to neurodegeneration (Barker et al. 2021). RUNX1 was identified as a prioritized gene implicated in the early stages of AD from another GWAS study (Yaghoobi and Malekpour, 2024). ITPR2 could lead to calcium toxicity in human neurons related to AD (Kawalia et al. 2017). FOXN3, a transcription factor, could regulate changes in gene expression in microglia during entorhinal cortex aging related to AD (Li et al. 2022). Microglia are the immune cells in the brain, and neuroinflammation can play a key role in the pathogenesis of AD and related neuro-diseases (Sragovich et al. 2021; Li et al. 2024). SORL1 could regulate APOE and CLU levels, two known AD-associated genes, in human neurons (Lee et al. 2023).

## 4.3 Insights into the Genetic Mechanism of AD

The statistically significant enriched KEGG pathways (with FDR values) based on the identified MCGs include Alzheimer's disease (< 0.01), Cardiac muscle contraction (0.02), Huntington's disease (0.03), Parkinson's disease (0.03), Non-alcoholic fatty liver disease (0.03), Thermogenesis (0.03), Apelin signaling pathway (0.03), cGMP-PKG signaling pathway (0.03), and Oxidative phosphorylation (0.03). Each pathway has an FDR < 0.05, suggesting that each pathway is unlikely to be identified by chance, thus increasing the confidence in its relevance to AD mechanisms. The pathway analysis has provided insights into the various aspects of genetic mechanism that govern AD. Disease pathways, including Alzheimer's Disease, Huntington's Disease, and Parkinson's Disease, highlight the shared molecular mechanisms across these different neurodegenerative diseases. The cardiac muscle contraction pathway and the cGMP-PKG signaling pathway (which regulates platelet function) suggest the importance of vascular factors. The non-alcoholic fatty liver disease pathway points towards the role of metabolic dysregulation in AD. The apelin signaling pathway is involved in several AD-related comorbidities, such as diabetes and cardiovascular disease. The thermogenesis and the oxidative phosphorylation pathways have revealed insights into the dysfunction in energy metabolism that may be implicated in AD.

## 4.4 Insights into Causal Biomarker Discovery

A central topic in biomedical studies and beyond is distinguishing true causal relationships from spurious correlations. This study defines "causal" as "the most probable" genetic biomarkers from a statistical perspective. Reverse-Gene-Finder is grounded in robust methodological innovation based on the key observation that genes with the highest probability of causing AD, defined as the MCGs, must have the highest probability of activating those neurons with the highest probability of causing AD, defined as the MCNs. Specifically, leveraging a genomic foundation model, it first actively performs interventions on genes (through masking out known AD-associated genes) to measure the causal effects of neurons on AD onset, strengthening the reliability of establishing causal rather than correlational relationships between MCNs and AD. It then uses backtracking to explicitly link MCNs back to MCTs, and their corresponding MCGs, making it possible to trace how MCNs are influenced by MCGs, thus better reflecting the causal relationships between MCGs and AD. The result, further validated via literature review and interpreted via gene enrichment analysis, is insightful because it offers a novel data-driven causal hypothesis generation paradigm, revealing putative causal genetic biomarkers of AD, which can accelerate further lab/clinical verification.

## 4.5 Interpretability, Generalizability, and Adaptability of the Reverse-Gene-Finder Technology

*Interpretability:* Our proposed Reverse-Gene-Finder technology enhances the interpretability of genomic foundation models for causal genetic biomarker identification. By perturbing known genes associated with AD and analyzing the causal impacts of neuron activations on model outputs, the localizations of MCNs have further revealed the complex relationships between the input layer and output mediated by MCNs (see Figure 2). Specifically, MCNs are primarily found in the first few and the intermediate layers, highlighting the importance of these middle layers in capturing subtle and complex genetic relationships for AD prediction. While some MCNs are located at the top or the bottom positions, the presence of MCNs closer to the middle positions indicates that the model can further extract complex gene expression patterns to reflect the genetic interactions collectively affecting AD, capturing the non-linear and multi-faceted nature of the AD genetic architecture (Neuner, Julia, and Goate 2020). Moreover, backtracking from MCNs to the input layer has made it possible to identify previously unknown genetic biomarkers that can cause AD based on their gene representations. This backtracking method differs from previous neural network-based gene-finding approaches, which operate from the input layer to the output and use gradient-based feature attribution for biomarker identification (Beebe-Wang et al. 2021; Zhang et al. 2023; Kassani et al. 2022). The proposed causal tracing and backtracking paradigm offers a novel interpretable approach for AI-driven biomarker identification.

*Generalizability:* Our proposed Reverse-Gene-Finder technology demonstrates its generalizability by identifying the MCGs relevant to multiple aspects of AD and implicated in broader neurodegenerative contexts, as supported by previous literature (see Section 4.2). Specifically, MCGs such as SORL1 can regulate APOE and CLU, two known AD-associated genes. Some MCGs have broad functional relevance to AD, such as PLXDC2 linked to amyloid beta and tau, MT-CO3/-CO1 related to oxidative stress contributing to AD, and MEF2A related to neuroprotection and neuroplasticity. Some MCGs, such as DOCK4 and FOXN3, are associated with other neurodegenerative disorders or involved in shared neuro-degeneration processes such as neuro-inflammation, highlighting cross-disease relevance and broader implications. The generalizability of the results is further reinforced by pathway analysis (see Section 4.3). By identifying pathways that link AD with other neurodegenerative diseases, vascular health, metabolic dysregulation, and energy metabolism, our proposed technology demonstrates the ability to provide a more comprehensive understanding of AD pathogenesis that can be generalized to broader contexts, and the potential for understanding the complex mechanisms underlying AD and related neurodegenerative diseases.

*Adaptability:* Our proposed Reverse-Gene-Finder mechanism is not limited to AD genetic biomarker identification. Identifying the causal genetic biomarkers is critical for early disease diagnosis and intervention (Goddard et al. 2016). The underlying principle of uncovering causal genetic biomarkers through causal tracing and backtracking is adaptable to other disease scenarios where genetic factors play a significant role. This is also a neural model-agnostic open-box transparent approach, which can be adapted to other neural architectures with a layered structure, such as language models, where interpretability has increasingly become critical for understanding model decision-making (Singh et al. 2024). In addition, the proposed technology can be adapted to a wide range of AI application domains beyond genetic biomarker identification, wherever understanding of the causal relationships across different input markers and discovery of new input markers are needed, such as linguistic-based disease diagnosis (Eyigoz et al. 2020).

## 5. Future Work

While the current work focuses on identifying novel causal biomarkers, integrating these biomarkers into predictive models to enhance AD prediction can be done in the future.

Future work can also extend the proposed Reverse-Gene-Finder technology to other AD biomarker identification scenarios. For instance, by backtracking language models, this technology has the potential to uncover new linguistic markers and refine the understanding of existing ones, further enhancing speech-based AD diagnostics (Eyigoz et al. 2020).

Moreover, future work can extend the proposed Reverse-Gene-Finder technology to identify causal biomarkers for various neurodegenerative diseases and other diseases. This technology can provide deeper insights into the mechanisms that predict disease onset by leveraging causal biomarkers identified through causal tracing and backtracking techniques. This deeper understanding could lead to the discovery of novel biomarkers, which could inform the development of more accurate diagnostic tools and targeted therapies for a broad spectrum of neurodegenerative conditions.

Finally, the robustness of the proposed Reverse-Gene-Finder technology in different settings can be examined by identifying causal biomarkers across different datasets and model architectures. Future work can investigate other types of biomedical foundation models, e.g., GPT-like models (Cui et al. 2024), and data modalities, e.g., single-cell imaging data (Yang et al. 2021) and multi-omics data (Efremova and Teichmann 2020) and develop a multi-modal approach for AD biomarker identification beyond genetics.

## 6. Conclusion

This study introduces a novel Reverse-Gene-Finder technology that leverages large genomic foundation models to identify previously unknown causal genetic biomarkers for AD. Our Reverse-Gene-Finder technology comprises three key innovations. Firstly, we exploit the observation that genes with the highest probability of causing AD, defined as the most causal genes (MCGs), must have the highest probability of activating those neurons with the highest probability of causing AD, defined as the most causal neurons (MCNs). Secondly, we utilize a gene token representation at the input layer to allow each gene (known or novel to AD) to be represented as a discrete and unique entity in the input space. Lastly, in contrast to the existing neural network architectures, which track neuron activations from the input layer to the output layer in a feed-forward manner, we develop an innovative backtracking method to track backwards from the MCNs to the input layer, identifying the MCTs and the corresponding MCGs. This approach offers new insights into the genetic mechanisms of AD. It can also be adapted to other diseases, providing new insights into the underlying mechanisms driving these diseases.

# Ethical Statement

This study has the potential to make a positive impact on public health by advancing our understanding of complex diseases and opening new possibilities for personalized medicine. The societal impacts of this study could be significant. If the proposed Reverse-Gene-Finder technology successfully identifies previously unknown genetic biomarkers for AD, it could lead to a better understanding of the genetic mechanisms behind the disease. These findings could potentially lead to earlier and more accurate diagnosis of AD, as well as the development of targeted healthcare interventions and treatments. Furthermore, the proposed technology could be adapted to study other diseases, leading to new insights and potential breakthroughs in various areas of healthcare. However, given that most medical datasets are not representative, the data-driven findings may be biased. Future work can investigate debiasing and fairness techniques to understand and mitigate potential biases.

# Acknowledgments

This work was supported in part by the United States National Academy of Medicine Healthy Longevity Catalyst Award (Grant No. HLCA/E-705/24), administered by the Research Grants Council of Hong Kong, awarded to V.O.K.L. and J.C.K.L, and by The Hong Kong University Seed Funding for Collaborative Research 2023 (Grant No. 109000447), awarded to V.O.K.L. and J.C.K.L.

# References

Alzheimer's Association. 2024. Alzheimer's Disease Facts and Figures. https://www.alz.org/alzheimers-dementia/facts-figures. Accessed: 2024-08-15.

Barker, S. J.; Raju, R. M.; Milman, N. E.; Wang, J.; Davila-Velderrain, J.; Gunter-Rahman, F.; Parro, C. C.; Bozzelli, P. L.; Abdurrob, F.; Abdelaal, K.; et al. 2021. MEF2 is a key regulator of cognitive potential and confers resilience to neurodegeneration. *Science Translational Medicine*, 13(618): eabd7695.

Bartolo, N. D.; Mortimer, N.; Manter, M. A.; Sanchez, N.; Riley, M.; O'Malley, T. T.; and Hooker, J. M. 2022. Identification and prioritization of PET neuroimaging targets for microglial phenotypes associated with microglial activity in Alzheimer's disease. *ACS Chemical Neuroscience*, 13(24): 3641–3660.

Beebe-Wang, N.; Celik, S.; Weinberger, E.; Sturmfels, P.; De Jager, P. L.; Mostafavi, S.; and Lee, S.-I. 2021. Unified AI framework to uncover deep interrelationships between gene expression and Alzheimer's disease neuropathologies. *Nature Communications*, 12(1): 5369.

Bennett, D. A.; Buchman, A. S.; Boyle, P. A.; Barnes, L. L.; Wilson, R. S.; and Schneider, J. A. 2018. Religious orders study and rush memory and aging project. *Journal of Alzheimer's Disease*, 64(s1): S161–S189.

Bertram, L.; McQueen, M. B.; Mullin, K.; Blacker, D.; and Tanzi, R. E. 2007. Systematic meta-analyses of Alzheimer disease genetic association studies: the AlzGene database. *Nature Genetics*, 39(1): 17–23.

Chung, J.; Wang, X.; Maruyama, T.; Ma, Y.; Zhang, X.; Mez, J.; Sherva, R.; Takeyama, H.; Lunetta, K. L.; Farrer, L. A.; et al. 2018. Genome-wide association study of Alzheimer's disease endophenotypes at prediagnosis stages. *Alzheimer's & Dementia*, 14(5): 623–633.

Cui, H.; Wang, C.; Maan, H.; Pang, K.; Luo, F.; Duan, N.; and Wang, B. 2024. scGPT: toward building a foundation model for single-cell multi-omics using generative AI. *Nature Methods*, 21(8): 1470–1480.

Downey, J.; Lam, J.C.K.; Li, V.O.K.; and Gozes, I. 2022. Somatic mutations and Alzheimer's disease. *Journal of Alzheimer's Disease*, 90(2): 475–493.

Efremova, M.; and Teichmann, S. A. 2020. Computational methods for single-cell omics across modalities. *Nature Methods*, 17(1): 14–17.

Escott-Price, V.; Shoai, M.; Pither, R.; Williams, J.; and Hardy, J. 2017. Polygenic score prediction captures nearly all common genetic risk for Alzheimer's disease. *Neurobiology of Aging*, 49: 214–e7.

Eyigoz, E.; Mathur, S.; Santamaria, M.; Cecchi, G.; and Naylor, M. 2020. Linguistic markers predict onset of Alzheimer's disease. *EClinicalMedicine*, 28: 100583.

Farrer, L. A.; Cupples, L. A.; Haines, J. L.; Hyman, B.; Kukull, W. A.; Mayeux, R.; Myers, R. H.; Pericak-Vance, M. A.; Risch, N.; and Van Duijn, C. M. 1997. Effects of age, sex, and ethnicity on the association between apolipoprotein E genotype and Alzheimer disease: a meta-analysis. *JAMA*, 278(16): 1349–1356.

Goddard, M.; Kemper, K.; MacLeod, I.; Chamberlain, A.; and Hayes, B. 2016. Genetics of complex traits: prediction of phenotype, identification of causal polymorphisms and genetic architecture. *Proceedings of the Royal Society B: Biological Sciences*, 283(1835): 20160569.

Jo, T.; Nho, K.; Bice, P.; Saykin, A. J.; and Initiative, A. D. N. 2022. Deep learning-based identification of genetic variants: application to Alzheimer's disease classification. *Briefings in Bioinformatics*, 23(2): bbac022.

Kassani, P. H.; Lu, F.; Le Guen, Y.; Belloy, M. E.; and He, Z. 2022. Deep neural networks with controlled variable selection for the identification of putative causal genetic variants. *Nature Machine Intelligence*, 4(9): 761–771.

Kawalia, S. B.; Raschka, T.; Naz, M.; de Matos Simoes, R.; Senger, P.; and Hofmann-Apitius, M. 2017. Analytical strategy to prioritize Alzheimer's disease candidate genes in gene regulatory networks using public expression data. *Journal of Alzheimer's Disease*, 59(4): 1237–1254.

Lee, H.; Aylward, A. J.; Pearse, R. V.; Lish, A. M.; Hsieh, Y.-C.; Augur, Z. M.; Benoit, C. R.; Chou, V.; Knupp, A.; Pan, C.; et al. 2023. Cell-type-specific regulation of APOE and CLU levels in human neurons by the Alzheimer's disease risk gene SORL1. *Cell Reports*, 42(8).

Lee, T.; and Lee, H. 2020. Prediction of Alzheimer's disease using blood gene expression data. *Scientific Reports*, 10(1): 3485.

Li, M.-L.; Wu, S.-H.; Song, B.; Yang, J.; Fan, L.-Y.; Yang, Y.; Wang, Y.-C.; Yang, J.-H.; and Xu, Y. 2022. Single-cell

analysis reveals transcriptomic reprogramming in aging primate entorhinal cortex and the relevance with Alzheimer's disease. *Aging Cell*, 21(11): e13723.

Li, V.O.K.; Han, Y.; Kaistha, T.; Zhang, Q.; Downey, J.; Gozes, I.; and Lam, J.C.K. 2024. DeepDrug: An Expert-led Domain-specific AI-Driven Drug-Repurposing Mechanism for Selecting the Lead Combination of Drugs for Alzheimer's Disease. medRxiv: 2024.07.06.24309990.

Liaw, R.; Liang, E.; Nishihara, R.; Moritz, P.; Gonzalez, J. E.; and Stoica, I. 2018. Tune: A research platform for distributed model selection and training. arXiv:1807.05118.

Lunnon, K.; Keohane, A.; Pidsley, R.; Newhouse, S.; Riddoch-Contreras, J.; Thubron, E. B.; Devall, M.; Soininen, H.; Kłoszewska, I.; Mecocci, P.; et al. 2017. Mitochondrial genes are altered in blood early in Alzheimer's disease. *Neurobiology of Aging*, 53: 36–47.

Luo, H.; and Specia, L. 2024. From Understanding to Utilization: A Survey on Explainability for Large Language Models. arXiv:2401.12874.

Mathys, H.; Davila-Velderrain, J.; Peng, Z.; Gao, F.; Mohammadi, S.; Young, J. Z.; Menon, M.; He, L.; Abdurrob, F.; Jiang, X.; et al. 2019. Single-cell transcriptomic analysis of Alzheimer's disease. *Nature*, 570(7761): 332–337.

Meng, K.; Bau, D.; Andonian, A.; and Belinkov, Y. 2022. Locating and editing factual associations in GPT. *Advances in Neural Information Processing Systems*, 35: 17359–17372.

Migliore, L.; and Coppede, F. 2022. Gene–environment interactions in Alzheimer disease: the emerging role of epigenetics. *Nature Reviews Neurology*, 18(11): 643–660.

Neuner, S. M.; Julia, T.; and Goate, A. M. 2020. Genetic architecture of Alzheimer's disease. *Neurobiology of Disease*, 143: 104976.

Park, C.; Ha, J.; and Park, S. 2020. Prediction of Alzheimer's disease based on deep neural network by integrating gene expression and DNA methylation dataset. *Expert Systems with Applications*, 140: 112873.

Rahimzadeh, N.; Srinivasan, S. S.; Zhang, J.; and Swarup, V. 2024. Gene networks and systems biology in Alzheimer's disease: Insights from multi-omics approaches. *Alzheimer's & Dementia*, 20(5): 3587–3605.

Shi, L. 2013. Dock protein family in brain development and neurological disease. *Communicative & Integrative Biology*, 6(6): e26839.

Singh, C.; Inala, J. P.; Galley, M.; Caruana, R.; and Gao, J. 2024. Rethinking interpretability in the era of large language models. arXiv:2402.01761.

Sragovich, S.; Gershovits, M.; Lam, J.C.K.; Li, V.O.K.; and Gozes, I. 2021. Putative blood somatic mutations in post- traumatic stress disorder-symptomatic soldiers: High impact of cytoskeletal and inflammatory proteins. *Journal of Alzheimer's Disease*, 79(4): 1723–1734.

Srivastava, A.; Barth, E.; Ermolaeva, M. A.; Guenther, M.; Frahm, C.; Marz, M.; and Witte, O. W. 2020. Tissue-specific gene expression changes are associated with aging in mice. *Genomics, Proteomics and Bioinformatics*, 18(4): 430–442.

Sun, N.; Victor, M. B.; Park, Y. P.; Xiong, X.; Scannail, A. N.; Leary, N.; Prosper, S.; Viswanathan, S.; Luna, X.; Boix, C. A.; et al. 2023. Human microglial state dynamics in Alzheimer's disease progression. *Cell*, 186(20): 4386–4403.

Szklarczyk, D.; Gable, A. L.; Lyon, D.; Junge, A.; Wyder, S.; Huerta-Cepas, J.; Simonovic, M.; Doncheva, N. T.; Morris, J. H.; Bork, P.; et al. 2019. STRING v11: protein–protein association networks with increased coverage, supporting functional discovery in genome-wide experimental datasets. *Nucleic Acids Research*, 47(D1): D607–D613.

Theodoris, C. V.; Xiao, L.; Chopra, A.; Chaffin, M. D.; Al Sayed, Z. R.; Hill, M. C.; Mantineo, H.; Brydon, E. M.; Zeng, Z.; Liu, X. S.; et al. 2023. Transfer learning enables predictions in network biology. *Nature*, 618(7965): 616–624.

Wang, J.; Ma, A.; Chang, Y.; Gong, J.; Jiang, Y.; Qi, R.; Wang, C.; Fu, H.; Ma, Q.; and Xu, D. 2021. scGNN is a novel graph neural network framework for single-cell RNA-Seq analyses. *Nature Communications*, 12(1): 1882.

Wang, Q.; Chen, K.; Su, Y.; Reiman, E. M.; Dudley, J. T.; and Readhead, B. 2022. Deep learning-based brain transcriptomic signatures associated with the neuropathological and clinical severity of Alzheimer's disease. *Brain Communications*, 4(1): fcab293.

Yaghoobi, A.; and Malekpour, S. A. 2024. Unraveling the genetic architecture of blood unfolded p-53 among non-demented elderlies: novel candidate genes for early Alzheimer's disease. *BMC Genomics*, 25(1): 440.

Yang, K. D.; Belyaeva, A.; Venkatachalapathy, S.; Damodaran, K.; Katcoff, A.; Radhakrishnan, A.; Shivashankar, G.; and Uhler, C. 2021. Multi-domain translation between single-cell imaging and sequencing data using autoencoders. *Nature Communications*, 12(1): 31.

Zhang, Q.; Han, Y.; Lam, J.C.K.; Bai, R.; Gozes, I.; and Li, V.O.K. 2023. An Expert-guided Hierarchical Graph Attention Network for Post-traumatic Stress Disorder Highly-associative Genetic Biomarkers Identification. medRxiv: 2023.01.30.23285175.